\documentclass[11pt]{article}

\usepackage[preprint]{acl}

\usepackage{times}
\usepackage{latexsym}

\usepackage[T1]{fontenc}

\usepackage[utf8]{inputenc}

\usepackage{microtype}

\usepackage{inconsolata}

\usepackage{graphicx}

\usepackage{amsmath,amssymb}
\usepackage{subcaption}
\usepackage{enumitem}
\usepackage{booktabs}
\usepackage{multirow}
\usepackage{xcolor}
\usepackage{makecell}

\definecolor{BrickRed}{HTML}{B6321C}

\newcommand\PaperTitle{Cartridges at Scale: Training Modular \\ KV Caches over Large Document Collections}

\title{\PaperTitle}

\author{
  Momchil Hardalov \quad Gonzalo Iglesias \quad Adri\`a de Gispert \\
  Amazon AGI  \\
  \texttt{\{momchilh, gjii, agispert\}@amazon.com}
}

\begin{document}
\maketitle
\begin{abstract}

Large Language Models can reason over long contexts, yet prefilling millions of tokens is wasteful as much of the content remains static across queries. Cartridges~\citep{eyuboglu2025cartridges} address this by distilling document collections into reusable key-value (KV) caches that eliminate prefilling while preserving accuracy. A critical limitation of this approach is that cartridges are monolithic and non-compositional: encoding an entire collection into a single KV block does not scale, and naively mixing cartridges trained in isolation collapses performance to near chance. We introduce \emph{Cartridges at Scale} (CAS), a training framework for scalable multi-cartridge learning with dynamic distractor mixing and a memory-efficient budget manager that rotates hundreds of per-document cartridges between GPU and persistent storage. Our approach scales to collections exceeding a million tokens, improving over a monolithic cartridge by 10--31 points at comparable token budgets. Oracle cartridge accuracy falls within 2--6 points of full in-context learning even at high compression. When paired with retrieval for cartridge selection, CAS matches or exceeds conventional RAG accuracy while consuming 3--4$\times$ fewer prompt tokens.

\end{abstract}

\section{Introduction}

\begin{figure}[t]
    \centering
    \includegraphics[width=0.95\columnwidth]{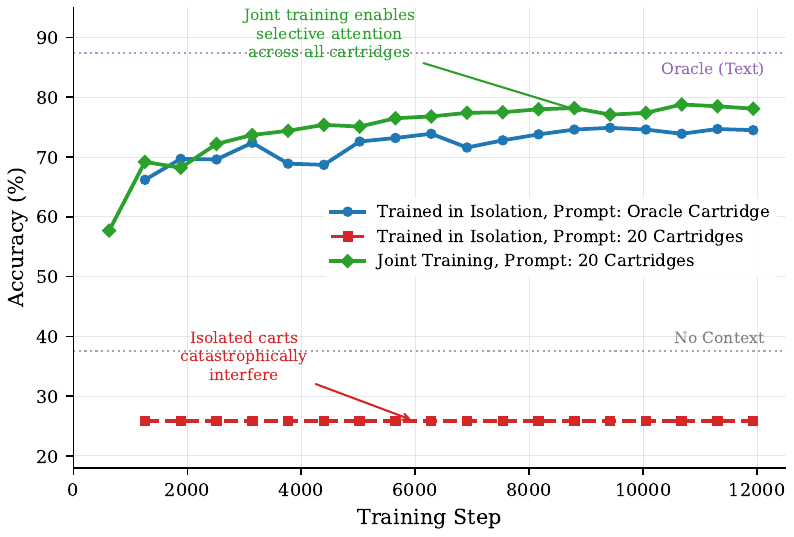}
    \caption{%
        Performance of per-document trained Cartridges on LongHealth~(\citet{adams2024longhealth}, docs: 20 patient notes, $\sim$232K tokens). \emph{Oracle} -- only the patient Cartridge is presented in the prompt; \emph{Full} -- all cartridges are presented. Details in \S~\ref{sec:experimenta_results}, \emph{Training Regime}.
    }
    \label{fig:isolation_vs_act20}
\end{figure}

Large language models (LLMs) can reason over information in their context window~\citep{snell2025scaling, jaech2024openai, openai2025, guo2025deepseek, comanici2025gemini, anthropic2025claude4, agarwal2025gpt}, however this capability comes at a cost that scales with context length---both in latency (prefill time) and memory (KV cache size).
For applications requiring persistent access to document collections---such as enterprise knowledge bases, patient records, and legal corpora---repeatedly processing the same documents for each query is prohibitively expensive.

Retrieval-augmented generation (RAG,~\citet{lewis2020retrieval,guu2020retrieval}) mitigates this by retrieving only the most relevant text chunks at query time, but each retrieved chunk is a \emph{fragment} of the original document, forcing the model to reason over incomplete information.

Prior work has proposed reducing KV-cache size via token eviction or merging~\cite{zhang2023h2o,kang2024gear,liu2026retrievalattention}, or compressing text into soft~\cite{mu2023gist,cheng2024xrag,louis-etal-2025-pisco} or discrete~\cite{jiang-etal-2023-llmlingua,pan-etal-2024-llmlingua} token representations. These approaches often require expensive runtime prefills, rely on a separate encoder, or lose information from long documents~\cite{lajewska-etal-2025-understanding}.

More recently, researchers have explored encoding documents into compact, reusable representations that can be loaded into the model KV cache at inference time without additional computational cost~\cite{devoto2025expected,kim2026kvzip,zweiger2026fast}.
\emph{Cartridges}~\cite{eyuboglu2025cartridges} proposed training a small set of key-value vectors per document collection via context distillation, achieving $10$--$100\times$ compression while preserving the model's ability to answer questions about the encoded content. However, their approach relies on a single monolithic cartridge per collection, which does not scale to real-world document collections: loading all documents into one cache bloats the prefix with irrelevant tokens, while information-dense content (e.g., financial tables, clinical notes, technical specifications) cannot be highly compressed without loss.
Moreover, a single monolithic cartridge is inflexible to future changes: updating or adding a single document requires re-encoding the entire KV cache. The natural alternative---one cartridge per document---exposes a more fundamental flaw: as Figure~\ref{fig:isolation_vs_act20} shows, \emph{mixing cartridges trained in isolation leads to model collapse}---each cartridge performs well in the \textcolor{blue!70!black}{oracle setting} (one correct cartridge loaded), but loading \textcolor{red!70!black}{all cartridges} simultaneously drops performance to near-chance, as the model never learned to selectively attend across independently trained KV prefixes.

In this work, we introduce CAS, a framework for training and deploying cartridges at scale over collections of hundreds of documents and millions of tokens. CAS proposes a training regime based on dynamic cartridge \textcolor{green!50!black}{rotation and mixing} that closes the performance gap of isolated training (Fig.~\ref{fig:isolation_vs_act20}) while remaining feasible within current GPU memory constraints. We further show that CAS integrates naturally with dense retrieval for cartridge selection, enabling a practical deployment path that matches or exceeds text RAG efficiency. Our contributions are as follows:

\begin{itemize}[leftmargin=*,nosep]
\item We show that independently trained cartridges suffer catastrophic interference at inference time, collapsing to near-chance performance, and that joint training with distractor cartridges removes this degradation.

\item We propose CAS, a scalable training framework with GPU $\leftrightarrow$ persistent storage swapping, optimizer offloading, and prioritized rotation, enabling joint training of hundreds of cartridges within a fixed GPU memory budget.

\item We propose improved self-study data generation via proportional-to-length sampling and batched multi-question generation, reducing synthesis cost by up to $20\times$ while improving fact coverage.

\item We demonstrate that per-document cartridges consistently outperform a single monolithic cartridge by up to $30$ points across five diverse benchmarks, with compression tolerance ranging from near-lossless at $100\times$ for short technical documents to up to 5\% relative performance drop for table-heavy financial filings.

\item We show that Cartridge RAG matches or exceeds Text RAG accuracy at up to $4\times$ fewer prompt tokens, combining the efficiency of compressed KV caches with the selectivity of dense retrieval.
\end{itemize}

\begin{figure*}[t]
    \centering
    \subcaptionbox{%
        \textbf{Self-Study.}
        For each document $d_i$, we sample a list of questions using $M_Q$. Subsequently, the answers along with the logits from the target teacher $M_A$ are collected.
        \label{fig:multicart_data}%
    }[0.30\textwidth]{%
        \includegraphics[width=0.37\textwidth]{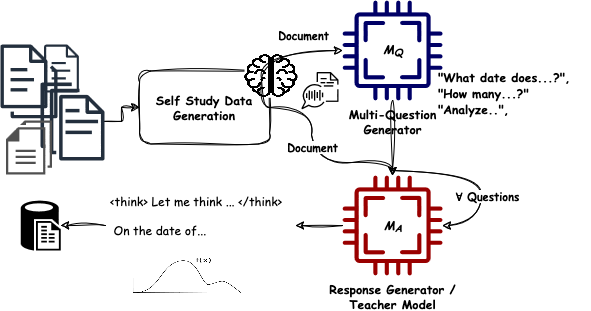}%
    }%
    \hfill
    \subcaptionbox{%
        \textbf{Training.}
        Each document $d_i$ is assigned a dedicated cartridge. A \emph{budget manager} keeps only $B$ cartridges on GPU and rotates the pool every $R$ steps. %
        \label{fig:multicart_training}%
    }[0.3\textwidth]{%
        \includegraphics[width=0.36\textwidth]{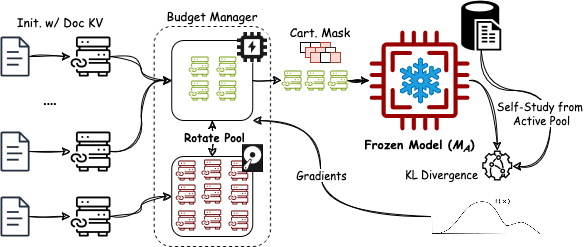}%
    }%
    \hfill
    \subcaptionbox{%
        \textbf{Inference.}
        A selection function activates $k$ cartridges from the store.
        The selected cartridges are concatenated and prepended to the query tokens before decoding.
        \label{fig:multicart_inference}%
    }[0.3\textwidth]{%
        \includegraphics[width=0.27\textwidth]{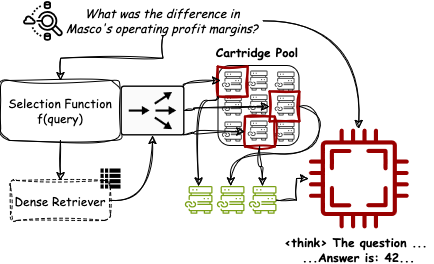}%
    }
    \caption{%
        Cartridges at Scale (CAS) End-to-End Pipeline.
    }
    \label{fig:cas}
\end{figure*}

\section{Cartridges at Scale (CAS) Training}

\paragraph{Background.} Our work builds on the Cartridge framework of \cite{eyuboglu2025cartridges}. A Cartridge $Z \in \mathbb{R}^{L \times p \times d \times 2}$ is a compact, trainable KV cache that encodes a context of $n_\mathcal{C}$ tokens as a fixed-size prefix of length $p \ll n_\mathcal{C}$ that is injected into the model's attention layers $L$ (of size $d$), yielding compression ratios of $10$--$100\times$ while integrating directly into existing inference servers.

The Cartridge 
is %
trained via a context-distillation objective using training data generated through \emph{Self-Study}: the corpus is chunked into subcorpora $\tilde{\mathbf{c}}$, and the LLM generates synthetic question--answer pairs about each chunk using diverse seed prompts~\citep{eyuboglu2025cartridges}. The loss minimizes the KL divergence between a \emph{teacher} (a model with the subcorpus in context) and a \emph{student} (the same model augmented with the Cartridge). All model parameters remain frozen---only the key-value vectors in $Z$ receive gradients.

\subsection{Training with Multiple Cartridges}

Although \cite{eyuboglu2025cartridges} discuss that independently trained Cartridges can be combined by concatenating their KV caches, this composition was never explicitly tested. Figure~\ref{fig:isolation_vs_act20} shows that mixing caches trained \emph{in isolation}---independently and without awareness of other Cartridges that may coexist at inference time---leads to model collapse. Additionally, naive scaling to very large Cartridge sizes is not feasible since training over large document collections has high memory requirements.\footnote{A 1K-token Qwen3-8B Cartridge takes $\sim551$\,MiB during backprop (500K tokens ${\sim}274$\,GiB). Gradients and Adam moments increase the Cartridges' memory footprint 4$\times$.} 

To address these limitations, we introduce CAS, a training and inference framework that supports the use of $N$ cartridges simultaneously (Figure~\ref{fig:cas}).

\paragraph{Mixed-visibility training.} Mixed-visibility training is key for learning KV representations that allow the frozen model to selectively attend to the relevant cartridge in the presence of distractors, directly optimizing for the multi-cartridge inference setting. To achieve this, we adapt the KL loss as follows: let $Z^* \in \{{Z_1,\ldots,Z_N}$\} be the relevant cartridge for example $\mathbf{x}$, $\mathcal{Z}\text{pool}$ the cartridge pool, $\mathcal{Z}\text{dist} = \mathcal{Z}\text{pool} \setminus {Z^*}$ the distractor candidates, and $k \sim \mathcal{U}(k_\text{min}, k_\text{max})$, then the loss becomes:

\begin{equation}
\tilde{Z} = \resizebox{.75\columnwidth}{!}{$\begin{cases}
Z^* & \text{w.p. } P_\text{iso} \\
Z^* \cup S,\; S \sim \mathcal{U}(\mathcal{Z}_\text{dist}, k) & \text{w.p. } 1 - P_\text{iso}
\end{cases}$} \label{eq:mixing}
\end{equation}

\begin{equation}
\resizebox{.88\columnwidth}{!}{$\mathcal{L}(Z) = \displaystyle\sum_{(\mathbf{x}, \tilde{\mathbf{c}}) \in \mathcal{D}} \sum_{i=1}^{|\mathbf{x}|} D_\text{KL}\!\left( \mathcal{F}(\cdot \mid \tilde{\mathbf{c}} \oplus \mathbf{x}_{<i}) \,\|\, \mathcal{F}_{\tilde{Z}}(\cdot \mid \mathbf{x}_{<i}) \right)$} \label{eq:cartridge_loss}
\end{equation}

Training with packing requires keeping all active cartridges in the prompt, while each packed sample sees a different subset via an attention mask.

\paragraph{Budget Manager} During training, we introduce a \emph{budget manager} that maintains a fixed GPU pool of $B \leq N$ cartridges, keeping the remaining $N - B$ cartridges offloaded to CPU memory or disk (Figure~\ref{fig:multicart_training}). At each step, the budget manager selects the active cartridges (i.e., those residing on GPU) and samples data from their training pool.

\paragraph{Pool rotation.}
Every $R$ optimizer steps, we rotate the active cartridges in the GPU pool: a fraction $\phi$ of the pool is evicted and replaced with new cartridges from persistent storage. Evicted cartridges receive no gradients until they are moved back to GPU. Setting $R = 1$ and $\phi = 1$ rotates the entire pool at every step, maximizing diversity but incurring the highest data-transfer overhead. Larger $R$ or smaller $\phi$ amortize this overhead by increasing the duration for which cartridges remain active, at the cost of slower turnover and reduced exposure to diverse cartridges during training. The rotation policy preferentially swaps in cartridges that have received the fewest optimizer steps, ensuring uniform document coverage.

To ensure the consistency of the RoPE embeddings~\cite{su2024roformer}, query token position IDs must be offset by the total number of active cartridge KV tokens, which occupy the prefix of the sequence. This offset is recomputed at each step based on the actual number of tokens contributed by the active cartridges.

\subsection{Optimization Dynamics}

\paragraph{Optimizer.} We use Adam~\cite{kingma2014adam} with fp32 precision, while we keep the weights in bfloat16, as this is crucial for numerical stability when scaling up the number of cartridges.

Instead of a fixed learning rate~\cite{eyuboglu2025cartridges}, we adopt a slow linear decay schedule with per-cartridge linear warmup to maintain a higher effective LR throughout training (see Appendix~\ref{appx:lr_decay}).

\paragraph{Per-cartridge LR warmup.}
When a cartridge is first swapped into the GPU pool, it has received zero optimizer steps and its gradients may be large relative to its current parameter values. Applying the global LR immediately can destabilize training. We therefore apply a per-cartridge linear warmup: a cartridge that has received $s$ optimizer steps uses an effective LR of $\eta_\text{eff} = \eta_t \cdot \min(1, \frac{s}{W_c})$, where $W_c$ is the per-cartridge warmup length. To avoid creating one optimizer param group per cartridge (which would prevent fused Adam kernels), we bucket the warmup scale into $N_b = 4$ discrete levels, yielding at most $N_b$ param groups per step.

\subsection{Improved Self-Study Data Synthesis}
\label{sec:data_synthesis}

The quality of synthetic training data is the primary driver of cartridge performance.
A detailed analysis on FinQA documents (Appendix~\ref{sec:synthesis_analysis}) reveals three key failures of the original procedure: (\emph{i})~\emph{uneven fact coverage}---since questions are generated without global awareness of the document, the synthesizer focuses on prose commentary rather than details such as figures in tables; (\emph{ii})~\emph{unbalanced document sampling}---facts are sampled uniformly across chunks, causing facts in longer, denser sections to be under-represented; and (\emph{iii})~\emph{low throughput}---one question per API call makes synthesis slow and expensive. To mitigate these limitations, we propose proportional-to-length sampling and batched multi-question generation (Figure~\ref{fig:multicart_data}).

\paragraph{Proportional-to-length sampling.}
We replace uniform random chunk sampling with a \emph{proportional-to-length} strategy ensuring balanced document coverage.
Each document $d_i$ in the collection is assigned a sampling weight $w_i = \frac{|d_i|}{\min_j |d_j|}$, so that longer documents---which are assumed to contain more facts---are sampled proportionally more often during synthesis.

\paragraph{Multi-question generation.}
We prompt the question-generation model to produce $n$ questions simultaneously (instead of doing $n$ LLM calls); we use $n{=}20$ and sample multiple times with temperature 0.6 (Figure~\ref{fig:multicart_data}).
The prompt instructs the model to vary the question style (factual recall, comparison, reasoning, detail-oriented) and to cover different facts, details, or aspects of the provided context.
Responses are returned as an array, parsed into individual questions and answered independently by the student model.

\paragraph{Decoupled question generation.}
We decouple the question-generation model $M_Q$ from the answer-generation $M_A$.
$M_Q$ is a larger, more capable model~\cite[GPT-OSS 120B]{agarwal2025gpt} that generates high-quality, diverse questions given the document context, whereas $M_A$ remains the target model whose KV cache we want to learn; 
it produces answers conditioned on the document---along with their token-level log-probabilities---as distillation targets for cartridge training (details in Appendix~\ref{appx:data_ablation}).

\subsection{Cartridge Initialization}

KV cache initialization quality matters for final Cartridge performance.
\citet{eyuboglu2025cartridges} show that initializing the cartridge using the KV cache from the first $p$ tokens of an arbitrary text outperforms random initialization, as it already projects the KVs within the model’s latent space.

\paragraph{Cart-specific initialization.} We propose a simple yet effective change to this procedure: when training over a collection of $N$ documents, each cartridge $Z_i$ is initialized from its own truncated document $d_i$ rather than from an arbitrary text. This ensures that the initial KV cache already encodes document-specific content, providing a warm start that significantly reduces the initial loss (over $\sim$50\% lower in our experiments), reduces the number of training steps required to reach convergence and leads to $\sim$10\% lower final loss.

We experimented with pooling-based compression strategies (mean/max pooling, top-$k$ norm selection, and sliding-window averaging) as alternatives for capturing more information from the original document, but none outperformed the cart-specific truncation strategy.

\begin{table*}[t!]
\centering
\resizebox{\textwidth}{!}{%
\small
\setlength{\tabcolsep}{3pt} %
\begin{tabular}{lrrrrrll}
\toprule
\textbf{Dataset} & \textbf{Docs} & \textbf{Questions} & \textbf{Qs/Doc} & \textbf{Avg.\ Tok.} & \textbf{Total Tok.} & \textbf{Task Type} & \textbf{Domain} \\
\midrule
LongHealth   & 20    & 400   & 20.0 & 11,700 &236K &  Multiple-choice (5-way) & Clinical patient records \\
QASPER       & 407   & 1,451 & 3.5 & 4,751  & 	665K &  Extract./free-form/yes-no & Full research papers \\
QuALITY      & 115   & 2,086 & 18.1 & 5,713  & 1.9M & Multiple-choice (4-way) & Fiction \& non-fiction narratives \\
T$^2$-RB/FinQA & 380 & 1,147 & 3.0 & 1,026 & 	392K & Math. Calculation & Corporate earnings reports w/ tables \\
TechQA       & 496   & 610   & 1.8 & 1,509  & 748K& Extractive & IBM IT support technotes \\
\bottomrule
\end{tabular}%
}
\caption{Dataset statistics. \textbf{Docs}: unique documents (one cartridge per document). \textbf{Qs/Doc}: average questions per document. \textbf{Avg.\ Tok.}: mean document length in tokens (Qwen3-8B tokenizer).}
\label{tab:datasets}
\end{table*}

\subsection{Inference}
\label{sec:inference}

At inference time, the model has access to a collection of $N$ trained cartridges $\{Z_1, \ldots, Z_N\}$, each encoding a distinct document $d_i$. When combining multiple cartridges we prepend their sequentially concatenated KV caches to the formatted prompt and question (Figure~\ref{fig:multicart_inference}), without stripping any special tokens from the processed inputs.

Because loading all $N$ cartridges simultaneously is both inefficient and infeasible for large collections (prefix length grows as $\mathcal{O}(N \cdot p)$), we select the $k$ most relevant cartridges for a given query $q$ before the forward pass. We evaluate two selection strategies: (\emph{i})~\emph{Oracle}, where the ground-truth cartridge is always loaded (an upper bound, not applicable in practice), and (\emph{ii})~\emph{Retrieved}, where the top-$k$ cartridges are selected by a dense retriever.

\section{Experimental Setup}

\paragraph{Models.} We use Qwen3-8B~\citep{yang2025qwen3} as the main model in our experiments (see Appendix~\ref{appx:hyperparameters} for training and inference details). 
Our RAG baseline uses dense retrieval via Amazon Bedrock Knowledge Bases (details in Appendix~\ref{sec:rag_details}).

\paragraph{Datasets.} We use these 5 datasets (Table~\ref{tab:datasets}):~\footnote{We filter out unanswerable questions and outlier documents that are significantly longer than others in the collection.}

\begin{itemize}[leftmargin=*,nosep]

\item \textbf{LongHealth} \cite{adams2024longhealth}: A clinical QA benchmark consisting of detailed fictional patient records with multiple-choice questions that test information extraction, negation understanding, and temporal sorting over long medical notes.
\item \textbf{QASPER} \cite{dasigi2021qasper}: An info-seeking QA dataset over full NLP research papers, where questions are written by readers who only saw the title and abstract. Answers include extractive spans, free-form text, and yes/no responses.
\item \textbf{QuALITY} \cite{pang2022quality}: A multiple-choice reading comprehension benchmark over long-form fiction and non-fiction narratives, where questions require careful reading and reasoning rather than surface-level pattern matching.
\item \textbf{T$^2$-RAGBench/FinQA} \cite{strich2026t2ragbench,chen2021finqa}: A decontextualized variant of FinQA designed for RAG evaluation, where questions about corporate earnings reports have been rewritten to be context-independent. The task requires numerical reasoning by constructing mathematical formulas over tables and text extracted from financial filings.
\item \textbf{TechQA} \cite{castelli2020techqa}: A technical support QA dataset drawn from IBM's Technote corpus, where questions require extracting precise solutions from IT documentation covering enterprise software and infrastructure issues.
\end{itemize}

\begin{table*}[t]
\centering
\resizebox{\textwidth}{!}{%
\small
\setlength{\tabcolsep}{5pt}
\begin{tabular}{lcccccc}
\toprule
\textbf{Method} & \textbf{Comp.} & \textbf{LongHealth (Acc.)} & \textbf{QuALITY} (Acc.) & \textbf{QASPER} (F1) & \textbf{FinQA} (EM) & \textbf{TechQA} (LLaaJ) \\
\midrule
No Context          & -- & 37.5 {\scriptsize$\pm$ 1.1} & 43.6 {\scriptsize$\pm$ 0.4} & 19.2 {\scriptsize$\pm$ 0.6} &  2.9 {\scriptsize$\pm$ 0.0} & 21.1 {\scriptsize$\pm$ 1.1} \\
Oracle Context      & $1\times$ & 87.4 {\scriptsize$\pm$ 0.8} & 82.5 {\scriptsize$\pm$ 0.3} & 56.7 {\scriptsize$\pm$ 0.4} & 66.8 {\scriptsize$\pm$ 2.7} & 74.7 {\scriptsize$\pm$ 0.9} \\
\midrule
\multirow{6}{*}{\makecell{CAS Training (Ours) \\ Single-Cartridge / Doc.}}
  & $2\times$    & \textbf{81.1} {\scriptsize$\pm$ 1.1}  & \textbf{78.6} {\scriptsize$\pm$ 0.9} & \textbf{54.9}$^{*}$ {\scriptsize$\pm$ 0.3} & \textbf{62.7} {\scriptsize$\pm$ 0.2} & \textbf{75.8} {\scriptsize$\pm$ 0.7} \\
  & $5\times$    & 79.9 {\scriptsize$\pm$ 0.8} & 78.5 {\scriptsize$\pm$ 0.9} & \textbf{54.9} {\scriptsize$\pm$ 0.6} & 57.9 {\scriptsize$\pm$ 0.4} & 75.6 {\scriptsize$\pm$ 0.7} \\
  & $10\times$   & 80.1 {\scriptsize$\pm$ 0.8} & 77.6 {\scriptsize$\pm$ 0.5} & 54.1 {\scriptsize$\pm$ 0.3} & 50.7 {\scriptsize$\pm$ 0.8} & 75.2 {\scriptsize$\pm$ 2.0} \\
  & $20\times$   & 78.9 {\scriptsize$\pm$ 1.0} & 77.6 {\scriptsize$\pm$ 0.6} & 53.9 {\scriptsize$\pm$ 0.6} & 45.0 {\scriptsize$\pm$ 0.9} & 75.0 {\scriptsize$\pm$ 0.4} \\
  & $50\times$   & 76.9 {\scriptsize$\pm$ 2.2} & 76.7 {\scriptsize$\pm$ 0.5} & 53.1 {\scriptsize$\pm$ 0.4} & 30.2 {\scriptsize$\pm$ 1.8} & 74.6 {\scriptsize$\pm$ 1.5} \\
  & $100\times$  & 77.3 {\scriptsize$\pm$ 1.4} & 76.4 {\scriptsize$\pm$ 0.8} & 53.1 {\scriptsize$\pm$ 0.3} & 23.0 {\scriptsize$\pm$ 0.3} & 74.7 {\scriptsize$\pm$ 0.9} \\
\bottomrule
\end{tabular}
}
\caption{No Context, Oracle Context, and Oracle Cartridge at selected compression ratios on Qwen3-8B. $\pm$ denotes std.\ dev.\ over 3 runs. \textbf{Bold} marks the best cartridge result per dataset. $^{*}$Trained with 3$\times$ due to memory constraints.}
\label{tab:compression_ablation}
\end{table*}

\paragraph{Retrieval.}
For our baseline RAG system, we use a chunk-based indexing of the documents, then we retrieve the relevant chunks based on the question, without reformulation (details in Appendix~\ref{sec:rag_details}). In the cartridge case, we use the RAG results as a proxy to retrieve relevant cartridges (see Figure~\ref{fig:multicart_inference}). In particular, we map each retrieved chunk to a document id, which in turn is associated with a cartridge. The final list contains the unique cartridge ids in the order of their retrieval. With 1,024-token chunks, LongHealth and QuALITY average ${\sim}2$ chunks per unique cartridge, while for TechQA and FinQA the ratio is closer to 1:1 (see Table~\ref{tab:unique_docs}). 

\paragraph{Evaluation}

Each dataset uses a task-specific system prompt during generation to elicit answers in the expected format. For FinQA, the model is instructed to produce structured mathematical formulas (e.g.,~\texttt{divide(subtract(x, y), y)}). For LongHealth and QuALITY, the model selects from multiple-choice options. For QASPER, the model provides brief extractive or yes/no answers. For TechQA, the model needs to produce concise factual answers.
The full model prompts that we use are provided in Appendix~\ref{sec:eval_prompts}. 

\section{Experimental Results}
\label{sec:experimenta_results}

\begin{table}[t]
\centering
\small
\resizebox{\columnwidth}{!}{%
\begin{tabular}{lcc}
\toprule
\textbf{Setting} & \textbf{Oracle (1 doc)} & \textbf{Full (20 docs)} \\
\midrule
Oracle Context              & 87.4 {\scriptsize$\pm$ 0.8} & --\,$^\dagger$ / 65.5 \\
\midrule
Train in Isolation  & 73.6 {\scriptsize$\pm$ 0.4} & \textcolor{BrickRed}{26.0 {\scriptsize$\pm$ 3.1}} \\
Train Jointly (25\% ex., 10 act.)    & \textbf{78.9} {\scriptsize$\pm$ 1.0} & 56.4 {\scriptsize$\pm$ 0.6} \\
Train Jointly (25\% ex., 20 act.)    & \textbf{79.0} {\scriptsize$\pm$ 1.4} & \textbf{77.8} {\scriptsize$\pm$ 0.6} \\
\bottomrule
\end{tabular}%
}
\caption{Effects of the training regime on inference. \emph{Oracle} -- only the gold cartridge; \emph{Full} -- all notes simultaneously. $\pm$ denotes std.\ dev.\ over 3 runs. $^\dagger$All notes exceeds the context; 65.5 uses 10 patients groups.}
\label{tab:isolation_ablation}
\end{table}

\paragraph{Training Regime.} We compare cartridge performance for LongHealth in two regimes: (\emph{i})~all cartridges are trained in isolation ($P_\text{iso}{=}1.0$), and (\emph{ii})~they are mixed in 1 out of every 4 examples ($P_\text{iso}{=}0.75$). In both regimes, we train with the same hyper-parameters---compression of $20\times$ (${\sim}585$ tokens/cart.), 80 epochs, and adding $k\sim\mathcal{U}(1, M)$ additional cartridges as distractors.

First, we focus on the oracle setting, where at inference time each question is given only the relevant cartridge in the KV prefix. In Table~\ref{tab:isolation_ablation} (\emph{Oracle}), the two joint training regimes achieve comparable performance—$78.9$ ($M=10$ active) and $79.0$ ($M=20$)—while isolated training remains at $73.6$. This suggests that single-cartridge quality benefits from exposure to distractors during training.

The gap widens drastically when we decode with all patient cartridges (\emph{Full}): joint training with 10 active cartridges degrades to $56.4$, while isolated training collapses to $26.0$. This shows that mixing cartridges trained in isolation leads to catastrophic interference, since the model never learned to distinguish between the distributions encoded in individual cartridges. Crucially, joint training with all 20 cartridges active nearly eliminates this degradation: the \emph{Full (20 docs)} score reaches $77.8$, only $1.2$ below its Oracle Cartridge result.

\begin{figure}[t]
\centering
\includegraphics[width=0.9\columnwidth]{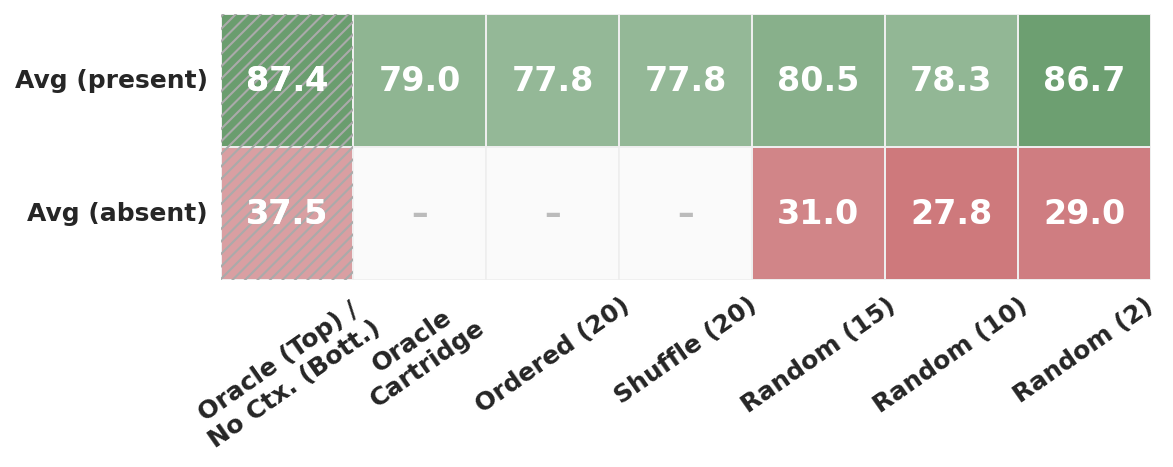}
\caption{Accuracy of \emph{Train Jointly (25\% ex., 20 act.)} on LongHealth varying the number of present cartridges.}
\label{fig:selectivity}
\end{figure}

\paragraph{Positional Invariance.} Figure~\ref{fig:selectivity} quantifies how the performance changes under different permutations of the cartridges. We see that shuffling the cartridge order has no effect ($77.8$ both for \emph{Ordered} and \emph{Shuffled}), demonstrating position-invariant attention over the KV prefix. When a random subset of $k$ cartridges is loaded, accuracy on questions whose cartridge \emph{is} present remains high ($78$--$87$), while accuracy on questions whose cartridge is \emph{absent} drops to $27$--$31$. This confirms that the model attends selectively to the correct cartridge rather than relying on a generic shared memory. Even though we back-propagate through all active cartridges simultaneously, there is a clear knowledge separation: each cartridge encodes document-specific information that the model can retrieve independently of the other cartridges in the prefix.

\paragraph{Scaling to Hundreds of Cartridges.}

Next, we scale up training to four additional datasets, each containing several hundred unique documents (Table~\ref{tab:datasets}). In this setting, we simulate realistic document collections with up to a million tokens.

Table~\ref{tab:compression_ablation} compares the model performance when consuming raw contexts vs. their compressed variants in individual cartridges. The \emph{No Context} row is the model performance without context, only using the question and the system prompt. We can see that all datasets perform poorly compared to the \emph{Oracle Context}, showing that the model is not able to solve the tasks parametrically. In the second part of the table, we show the best results at a given target compression ($2\times$ to $100\times$)\footnote{Each document is compressed to its length divided by the target compression.} for each dataset.

\begin{figure*}[t]
\centering
\includegraphics[width=\textwidth]{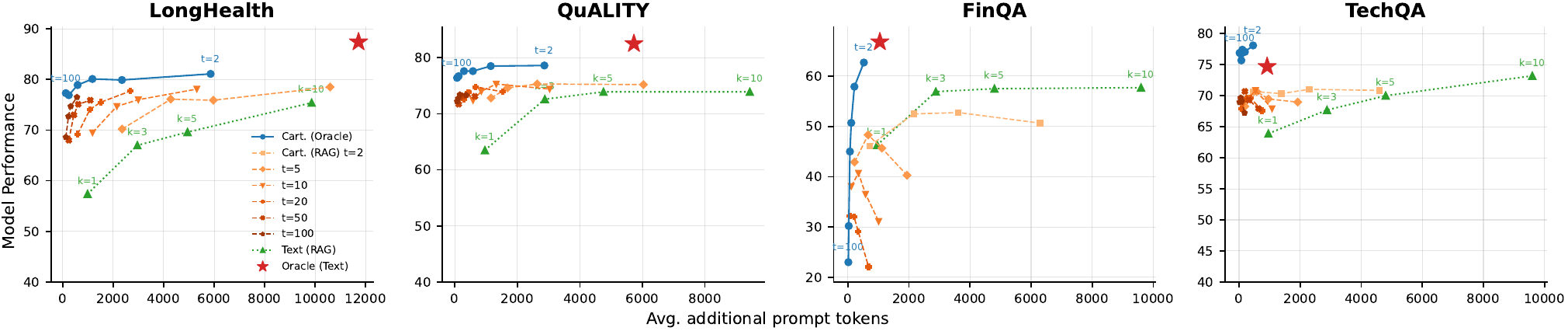}
\caption{Accuracy vs.\ additional prompt tokens trade-off between text and Cartridge RAG. The red star denotes the text oracle score; $t$: compression ratio; $k$: number of retrieved chunks.}
\label{fig:perf_vs_tokens}
\end{figure*}

We see the following patterns across datasets: (\emph{i})~the achievable compression depends strongly on document complexity: factually dense datasets and longer documents ({LongHealth}) do not tolerate high compression rates, while extraction tasks that do not require reasoning ({TechQA}) have no loss in performance even at $100\times$ compression; (\emph{ii})~the parametric knowledge of the model is stabilizing the compression -- {QuALITY} and {QASPER} contain books/papers that are included in many pre-training corpora~\cite{gao2020pile,weber2024redpajama} and have only a low drop at higher compressions, (\emph{iii})~dense structures in the text, such as tables, are hard to compress, i.e.,~{FinQA} drops 4-9 points at $2\times$-$5\times$ compression, similar to {LongHealth}.

\begin{table}[t]
\centering
\resizebox{\columnwidth}{!}{%
\small
\setlength{\tabcolsep}{4pt}
\begin{tabular}{lcccccc}
\toprule
\textbf{Token Budget} & \multicolumn{2}{c}{\textbf{LongHealth}} & \multicolumn{2}{c}{\textbf{QuALITY}} & \multicolumn{2}{c}{\textbf{FinQA}} \\
\cmidrule(lr){2-3} \cmidrule(lr){4-5} \cmidrule(lr){6-7}
& Compr. & Acc. & Compr. & Acc. & Compr. & EM \\
\midrule
4{,}096  & 57$\times$ & 67.2 & 160$\times$ & 62.1 & 96$\times$ & 6.9 \\
20{,}000 & 12$\times$ & 70.5 & 33$\times$  & 64.2 & 20$\times$ & 14.1 \\
28{,}672 & 8$\times$  & 72.6 & 23$\times$  & 65.2 & 14$\times$ & 13.0 \\
\midrule
\textit{Multi-cart.} & \textit{10$\times$} & \textit{80.1} & \textit{20$\times$} & \textit{77.6} & \textit{20$\times$} & \textit{45.0} \\
\bottomrule
\end{tabular}}
\caption{Performance of a monolith cartridge trained on all documents under varying budgets.}
\label{tab:single_cart}
\end{table}

\paragraph{Cartridges Granularity.}
We ablate whether the gains come from \emph{compression capacity} (the cartridge has $\geq p$ KV slots) or from \emph{per-document specialization} (each document has its own cartridge). We disentangle the two by training a single monolithic cartridge on the concatenation of all documents, with cartridge sizes of 4K, 20K, 28.7K tokens---the latter being roughly the full available KV budget given a $32$K context window (no YaRN~\citet{peng2024yarn}) minus prompt and generation overhead.

Despite allocating up to 28K tokens, the monolith baseline (Table~\ref{tab:single_cart}) achieves only $72.6$ on LongHealth, $65.2$ on QuALITY, and $14.1$ on FinQA. In contrast, per-document cartridges (\emph{Multi-cart.}) reach $80.1$ on LongHealth at $10\times$ compression (${\sim}1.2$K tok./doc, $23{.}6$K total), $78.6$ on QuALITY at $2\times$ (${\sim}2{,}9$K tok./doc), and $62.7$ on FinQA at $2\times$ (${\sim}524$ tok./doc). This large gap confirms that per-document specialization is critical. The FinQA results are particularly striking: 380 financial documents with dense numerical content cannot be meaningfully compressed into a single shared representation, as gradient updates from different documents destructively interfere. Nonetheless, dedicating a separate cartridge to each doc resolves this interference entirely in the oracle setting.

\paragraph{Cartridge RAG.} The \emph{Oracle} results assume the gold cartridge is always loaded, but in practice we often do not know which document or context is relevant. Moreover, performing inference with hundreds or thousands of cartridges in the prompt is inefficient and even infeasible. Here, we propose a hybrid approach combining standard dense retrieval with cartridge inference: given a query, we retrieve the top-$k$ document chunks, resolve the corresponding cartridge identifiers, and load only those cartridges into the KV prefix instead of the raw chunk text.
While \emph{Text RAG} appends $k$ chunks of ${\sim}1{,}024$ tokens each (consuming $k{\times}1\text{K}$ prompt tokens; see Appendix~\ref{sec:rag_details}), \emph{Cartridge RAG} loads the corresponding document cartridges at $t{\times}$ compression, e.g., at $t{=}20$, a $12$K-token document occupies only ${\sim}600$ KV tokens, representing the \emph{entire} document rather than a single chunk fragment.

Figure~\ref{fig:perf_vs_tokens} reports accuracy as a function of the additional prompt tokens consumed at inference.\footnote{We exclude QASPER from the RAG comparison; the data is not decontextualized (``What datasets do they use?'').} On LongHealth and QuALITY, where retrieval recall is high (${\geq}95\%$ at $k{=}10$; Appendix~\ref{sec:rag_recall}), \emph{Cartridge RAG} clearly dominates \emph{Text RAG}. On LongHealth, \emph{Cartridge RAG} at $t{=}20$, $k{=}10$ matches the best \emph{Text RAG} ($75.4$) using only $2{,}673$ tokens---$3.7\times$ fewer than Text RAG's $9{,}860$---and at $t{=}100$, $k{=}10$ still delivers $76.5$ at just $566$ tokens. On QuALITY, Cartridge RAG with $t{=}5$, $k{=}5$ reaches $75.3$, surpassing Text RAG ($73.9$) by $1.4$ at roughly half the token budget ($2{,}635$ vs.\ $4{,}754$). On FinQA, we observe a clear trend where loading more cartridges \emph{hurts}: at $t{=}2$, accuracy peaks at $k{=}5$ ($52.8$) and drops to $50.7$ at $k{=}10$, consistently across all compression ratios.
We hypothesize that the model becomes overwhelmed by the dense numerical content encoded in multiple financial-document cartridges, and that increasing the number of active cartridges during training (currently capped at 10) may alleviate this. On TechQA, \emph{Text RAG} slightly outperforms \emph{Cartridge RAG} ($73.2$ vs.\ $71.0$), likely because answers require extracting precise spans from the gold text---a setting where verbatim token access in raw chunks is more effective than compressed KV representations from multiple documents that bring noise. Still, in the TechQA case \emph{Cartridge RAG} remains more effective until retrieving more than 5 chunks.

\section{Discussion}

\paragraph{Cost Analysis.} The main cost of using cartridges lies in the offline training process which includes self-study data generation: sampling ~100-200 questions/doc. and doing one inference with the target model to generate the response\footnote{We are modeling the KL divergence which allows for training on partial sequences, unlike in cross-entropy loss.}. Depending on the size, complexity, and information density of documents, the training epochs can vary but often around 10 to 20 epochs are enough to reach 95\% of the best cartridge performance (Figure~\ref{fig:isolation_vs_act20}).

Inference costs are limited to loading the cartridges; there are no additional inference costs since we do not need any prefills. At inference, since prefill cost scales as the attention cost ($\mathcal{O}(n^2)$), a $10\times$ token reduction yields ${\sim}100\times$ fewer prefill FLOPs: Cartridge at $r{=}10$ uses ${\sim}1{,}200$ tokens vs.\ ${\sim}12{,}000$ for the text Oracle and ${\sim}9{,}860$ for Text RAG ($k{=}10$) on LongHealth.

Storage is another factor to consider. The KV cache footprint in BF16 per 1K tokens scales with model size: ${\sim}110$\,MiB for Qwen3-0.6B, ${\sim}141$\,MiB for 4B/8B, and ${\sim}250$\,MiB for 32B.
\paragraph{Document grouping.}
Our proposed CAS framework naturally extends to \emph{grouped cartridges} encoding multiple related documents (e.g.,~quarterly reports from the same company, chapters of the same book).
Grouping exploits shared vocabulary and background knowledge, allowing the token budget to focus on unique information that can help achieve even higher compression.
It also addresses the retrieval recall ceiling on homogeneous corpora: by clustering co-relevant documents, the retriever operates over a smaller, more discriminative set of groups, and a single retrieved group cartridge provides richer context than any single-document cartridge. We leave the systematic study of grouping strategies to future work.

\section{Related Work}

\paragraph{Soft Prompting.} These methods are variants of prefix-tuning methods~\citep{Lester:2021:EMNLP} that compress a given text into continuous representations~\cite{li-etal-2025-prompt}. Similar to KV cache compression, soft compression encodes long contexts into compact soft representations that can be prepended to queries, allowing for storing and caching. AutoCompressors~\citep{chevalier-etal-2023-adapting} recursively generate summary vectors; Gisting~\citep{bulatov2022rmt,mu2023gist} condenses prompts into virtual gist tokens; and other methods encode contexts into compact memory slots via LoRA-adapted or KV-based encoders~\citep{Ge:2024:ICLR,Li:2024:arXiva}. xRAG~\citep{cheng2024xrag} and PISCO~\citep{louis-etal-2025-pisco} take a modular approach, projecting retrieval embeddings or distilled representations into the LLM's input space while keeping the decoder frozen. These methods differ primarily in their training objectives (KL divergence vs. sequence-level distillation) and in which components remain trainable.

\paragraph{KV Cache Compression.} A natural way to reduce the context is to apply operations to its KV cache. This can happen either after or during the prefill phase by evicting non-impactful key-value pairs~\cite{zhang2023h2o,oren-etal-2024-transformers,ge2024model,tang2024quest,chari2025compactor,ancucki2025inferencetime} or merging them~\cite{wang2024modeltells,zhang2024cam,liu2025zsmergezeroshotkvcache,wan2025d2o}. SnapKV~\cite{li2024snapkv} and PyramidKV~\cite{cai2025pyramidkv} evict low-importance KV entries based on attention scores observed during prefill, with PyramidKV additionally allocating non-uniform per-layer budgets. KVzip~\cite{kim2026kvzip} selects tokens to retain via a learned context-reconstruction objective.
Attention Matching~\cite{zweiger2026fast} optimizes a compacted cache to reproduce the original attention distribution instead of reconstructing individual KV vectors. RetrievalAttention~\cite{liu2026retrievalattention} takes an orthogonal approach, offloading the full KV cache to the CPU and using approximate nearest-neighbor search to fetch the relevant entries.

We view many of these KV cache techniques as complementary to cartridges: compacted caches can serve as a more effective initialization point, which is then fine-tuned to the collection distribution; additionally, compaction can be applied to the trained cartridges because their representation is query-independent and may admit further compression for specific queries.

\section{Conclusions}
We present an end-to-end framework for training and serving multiple cartridges at a granularity as small as that of individual documents. Our framework addresses a fundamental flaw in the original monolithic Cartridges design, which prevents mixing independently trained cartridges, leading to collapse in performance at near chance levels. Experiments across five long-context datasets demonstrate that our approach scales to hundreds of cartridges over hundreds of thousands to millions of tokens, outperforming monolithic designs by 10–30 points at comparable computational budgets. We further propose improved initialization strategies and more efficient self-study generation, achieving superior accuracy and efficiency. Compression tolerance varies with document complexity ($100\times$ for TechQA, ${\leq}2\times$ for FinQA), reaching within 5\% of uncompressed baselines. Finally, we show that cartridges can be combined with retrieval, matching RAG performance while using 3--4$\times$ fewer tokens, though for information-dense tasks such as FinQA, text RAG maintains an edge.

\section*{Acknowledgments} 
We want to thank Rexhina Blloshmi, Felix Hieber, Bill Byrne, Hagen F{\"u}rstenau and Karan Gill for the time they took to support, discuss and improve this work. 

\section*{Limitations}

\paragraph{Cartridge KV position.}
All of our experiments prepend the trained KV cache to the prompt and query tokens, which is the natural setting for single-turn QA: the cartridge occupies the earliest positions and all subsequent tokens attend to it via standard causal masking.
In multi-turn scenarios, however, a cartridge may need to be loaded mid-conversation---for example, when a user references a new document after several turns of dialogue. Prepending the cartridge at that point would invalidate the previously computed KV entries, requiring a full recomputation of the conversation history.

\paragraph{Cartridge-specific retrieval.}
Our Cartridge RAG pipeline reuses a standard text-based dense retriever to select which cartridges to activate.
This is a pragmatic choice, but it means retrieval quality is bounded by how well the query matches the \emph{raw document text} rather than the compressed KV representation.
A dedicated cartridge retriever---one that learns to match queries directly against cartridge embeddings---could improve recall, particularly on corpora where text-based retrieval is unreliable.
Developing such a retriever is a natural next step, but requires either a learned projection from cartridge KV space to a query-compatible embedding space, or a contrastive training objective that aligns query representations with cartridge representations.

\paragraph{Language coverage.}
All five benchmarks used in this work are in English, and the base model (Qwen3-8B) is predominantly trained on English text.
It is unclear whether the self-study synthesis procedure, which relies on the model's own generative capabilities to produce training data, transfers equally well to lower-resource languages where the model's generation quality is weaker.
We do not evaluate cartridge performance on non-English corpora and cannot make claims about multilingual generalization.

\paragraph{Model scale.}
All experiments use a single model size. Larger models may achieve higher compression ratios at the same quality level, since their greater parametric capacity can compensate for more aggressive KV compression. Conversely, smaller models may require lower compression ratios to retain acceptable performance. Nonetheless, previous work~\cite{eyuboglu2025cartridges,zweiger2026fast} showed that cartridge performance scales across model sizes and model families.

\section*{Potential Risks}

Cartridges enable efficient, persistent encoding of arbitrary document collections into reusable KV caches. A malicious actor could exploit this to encode and serve disinformation, harmful instructions, or private data at scale. Additionally, documents encoded into cartridges may retain sensitive information in their KV representations in ways that are difficult to audit or redact---unlike raw text retrieval, it is unclear to what extent personal or confidential content can be extracted from a trained cartridge, raising data governance concerns in enterprise or medical deployments. We encourage practitioners to apply the same content-filtering, bias mitigation, and access-control measures to cartridge collections as they would to the underlying documents.

\bibliography{custom}

\clearpage

\appendix

\section{Hyperparameters}
\label{appx:hyperparameters}

Table~\ref{tab:hyperparams} summarizes the training hyper-parameters used across all datasets. All experiments use Qwen3-8B\footnote{\url{https://huggingface.co/Qwen/Qwen3-8B}} as the base model with BF16 weights and FP32 Adam optimizer states. Cartridge parameters are the only trainable parameters; all model weights remain frozen. We train using a cluster with NVidia H200 and B200 cloud GPUs.

We use Qwen3-8B as released on Hugging Face. The model card provided by the Qwen team documents training data, intended use, and limitations~\cite{yang2025qwen3}.

\begin{table*}[t]
\centering
\small
\begin{tabular}{lcc}
\toprule
\textbf{Hyperparameter} & \textbf{Value} & \textbf{Notes} \\
\midrule
\multicolumn{3}{l}{\emph{Optimization}} \\
Optimizer & Adam (FP32) & Moment tensors offloaded to NVMe for CPU-resident carts \\
Peak learning rate ($\eta_0$) & 0.05--0.1 & Dataset-dependent; 0.1 for LongHealth, 0.05 for others \\
LR schedule & Linear decay with warmup & \\
Warmup steps ($W$) & 200 & Global warmup \\
Final LR multiplier ($\alpha_f$) & 0.02 & $\eta_\text{min} = 0.02 \cdot \eta_0$ \\
Warmup minimum LR & 0.002 & \\
Per-cartridge warmup ($W_c$) & 20 steps & Applied when a cartridge first enters the GPU pool \\
Training epochs & 80 & \\
Global batch size & 128 & \\
Packed sequence length & 8{,}192 tokens & \\
\midrule
\multicolumn{3}{l}{\emph{Budget Manager \& Rotation}} \\
Cartridge budget ($B$) & 20 & Max cartridges on GPU simultaneously \\
Rotation interval ($R$) & 10 steps & Steps between pool rotations \\
Swap fraction ($\phi$) & 0.5 & Fraction of pool replaced per rotation \\
Prioritize least trained & Yes & Cartridges with fewest steps swapped in first \\
Empty cache interval & 5 rotations & \texttt{torch.cuda.empty\_cache()} frequency \\
\midrule
\multicolumn{3}{l}{\emph{Cartridge Masking}} \\
Isolation probability ($P_\text{iso}$) & 0.75 & Prob.\ of seeing only the relevant cartridge \\
Max active cartridges ($k_\text{max}$) & 10 & Max distractors visible per example \\
Min active cartridges ($k_\text{min}$) & 1 & \\
\midrule
\multicolumn{3}{l}{\emph{Initialization \& Padding}} \\
Init strategy & Cart-specific & Each cartridge initialized from its own document \\
Cache padding & Repeat & Tiles existing KV vectors if doc $<$ cart size \\
Num frozen tokens & 1 & BOS token frozen during training \\
\midrule
\multicolumn{3}{l}{\emph{Infrastructure}} \\
Hardware & NVIDIA H200 \& B200 & \\
Precision & BF16 (weights), FP32 (optimizer) & \\
Gradient checkpointing & Disabled & \\
\texttt{torch.compile} & Disabled & \\
\bottomrule
\end{tabular}
\caption{Training hyperparameters for the CAS framework. Values shown are the defaults used across all five datasets unless noted otherwise. The peak learning rate is the primary dataset-dependent hyperparameter.}
\label{tab:hyperparams}
\end{table*}

\paragraph{Dataset-specific variations.}
The total number of optimizer steps varies by dataset due to different corpus sizes: LongHealth (20 docs) trains for ${\sim}12{,}480$ steps, QuALITY (115 docs) for ${\sim}12{,}400$ steps, QASPER (407 docs) for ${\sim}14{,}160$ steps, and FinQA (380 docs) for ${\sim}14{,}160$ steps. The LR schedule \texttt{max\_steps} parameter is set accordingly. For the isolation ablation in Table~\ref{tab:isolation_ablation}, we use $P_\text{iso}{=}0.75$ for joint training and $P_\text{iso}{=}1.0$ for isolated training, with all 20 cartridges active simultaneously (no rotation needed for 20 cartridges on a H100 GPU).

\paragraph{Model Inference.}
For all experiments we evaluate the models with thinking mode enabled, temperature 0.6, and 2048 maximum completion tokens. We use a separate prompt for each dataset (see Appendix~\ref{sec:eval_prompts}), which we pass in the user turn. In the RAG condition, the retrieved context replaces the full document in the prompt; in the Oracle text condition, the full document text is provided. All reported results are averaged across at least 3 runs.

\section{Effective LR per Cartridge}
\label{appx:lr_decay}

\begin{figure}[t]
\centering
\includegraphics[width=\columnwidth]{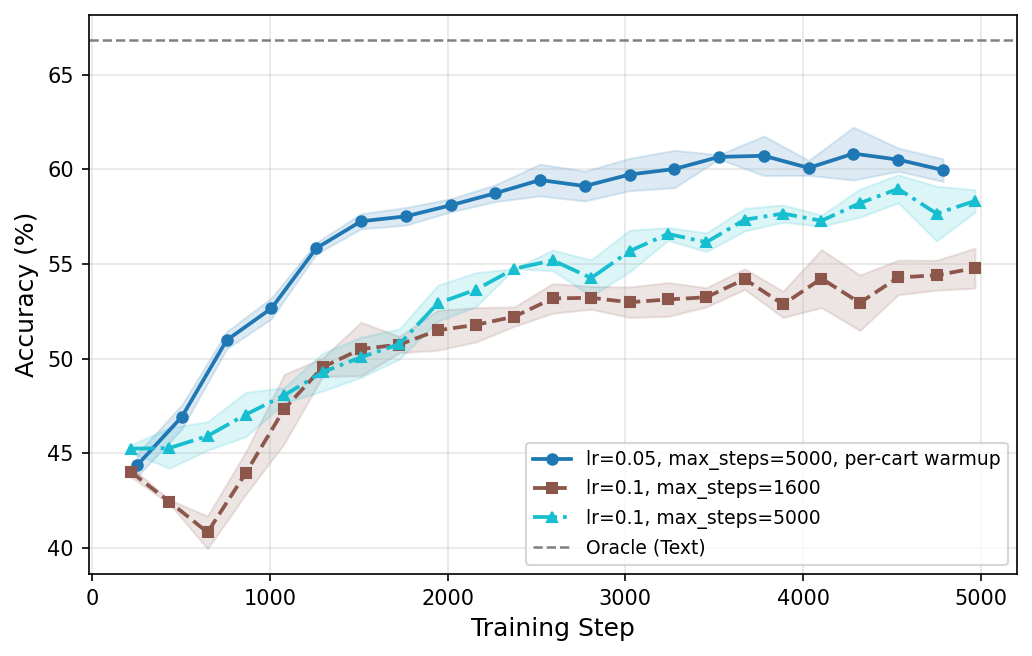}
\caption{LR schedule ablation on FinQA. The slow-decay schedule with $\texttt{max\_steps}{=}5000$ and per-cartridge warm-up is superior to $\texttt{max\_steps}{=}1600$.}
\label{fig:lr_decay_ablation}
\end{figure}

Because cartridge rotation means each cartridge is active for only a fraction of total training steps, the effective number of gradient updates per cartridge is much smaller than the global step count.
A standard short-decay schedule causes the learning rate to reach its minimum long before most cartridges have received sufficient updates, stalling their convergence.
We ablate this on FinQA by comparing a short-decay schedule ($\texttt{max\_steps}{=}1600$) against a slow-decay schedule ($\texttt{max\_steps}{=}5000$, $\alpha_f{=}0.02$) that maintains a higher effective learning rate throughout training (Figure~\ref{fig:lr_decay_ablation}).
The short-decay schedule reaches $54.8\%$; switching to slow decay improves this to $59.0\%$ (+4.2 points).
Adding per-cartridge warmup and prioritize-least-trained scheduling further pushes accuracy to $60.8\%$, as each cartridge receives a proper warmup phase when first loaded into the GPU budget.

\section{Impact of Self-Study Data Improvements}
\label{appx:data_ablation}

We ablate the cumulative effect of our self-study data improvements on FinQA using 10 single-doc cartridges of 512 tokens each. We compare the three data generation strategies:

\begin{itemize}[nosep]
\item \textbf{Original}: The baseline procedure from~\citet{eyuboglu2025cartridges}---one question per LLM call, using the target model (Qwen3-8B) for both question and answer generation, with uniform random chunk sampling.
\item \textbf{Decoupled $M_Q$, $n{=}20$}: Multi-question generation ($n{=}20$ questions per call) with a decoupled, more capable question model (GPT-OSS 120B) while keeping the target model as $M_A$. Same uniform sampling.
\item \textbf{Decoupled $M_Q$, $n{=}20$ + prop.\ to length}: Same as above, but replacing uniform sampling with proportional-to-length document sampling, ensuring longer documents receive proportionally more training examples.
\end{itemize}

\begin{table}[t]
\centering
\small
\begin{tabular}{lc}
\toprule
\textbf{Data Strategy} & \textbf{Accuracy (\%)} \\
\midrule
Original & 55.9 \\
Decoupled $M_Q$, $n{=}20$ & 62.7 \\
Decoupled $M_Q$, $n{=}20$ + prop.\ to length & \textbf{64.7} \\
\midrule
Oracle (Text) & 66.8 \\
\bottomrule
\end{tabular}
\caption{Impact of self-study data improvements on FinQA (10 cartridges, 512 tokens/cart).}
\label{tab:data_ablation}
\end{table}

The results (Table~\ref{tab:data_ablation}) show that decoupling the question model and batching 20 questions per call provides the largest single improvement (+6.8 points over the original), by generating more diverse and higher-quality questions that better cover the document's factual content. Adding proportional-to-length sampling contributes a further +2.0 points by ensuring that longer, information-dense documents receive training signal commensurate with their expected greater complexity. 

\section{Scoring Metrics}
\label{sec:metrics}

We use dataset-specific scoring metrics:

\begin{itemize}[nosep]
\item \textbf{T$^2$-RAGBench/FinQA}: Formula execution accuracy. The predicted formula is parsed and executed; the numerical result is compared to the gold answer with a relative tolerance of 1\%.
\item \textbf{LongHealth}: Fuzzy option matching. The predicted answer is compared against all five options using string similarity, and the closest match is selected.
\item \textbf{QASPER}: Average token F1 with multi-reference support, following the standard QASPER evaluation protocol. For each question, we compute the token-level F1 against all available reference answers and take the maximum.
\item \textbf{QuALITY}: Exact match on the extracted answer letter (A, B, C, or D).
\item \textbf{TechQA}: LLM-as-a-Judge. A separate judge model (DeepSeek-Distilled-Qwen-32B) evaluates whether the generated answer is factually equivalent to the reference answer. The judge outputs a JSON object with a justification and a binary grade. A score of 1 is assigned when the judge returns \texttt{"grade": "correct"}, and 0 otherwise. Table~\ref{tab:llmaaj-prompt} shows the exact judge prompt.

\begin{table*}[t]
\centering
\footnotesize
\begin{tabular}{@{}p{0.12\textwidth}p{0.85\textwidth}@{}}
\toprule
\textbf{Role} & \textbf{Prompt} \\
\midrule
System &
\texttt{You are to act as an impartial judge, evaluating whether an answer to a question matches a provided reference answer. Your task is to determine if the given answer is correct based on its factual equivalence to the reference answer, ignoring differences in punctuation and phrasing.} \\
& \\
& \texttt{When evaluating the answer, consider the following criteria:} \\
& \texttt{1. Factual equivalence: Does the answer convey the same information as the reference answer?} \\
& \texttt{2. Completeness: Does the answer address all parts of the question that the reference answer addresses?} \\
& \texttt{3. Accuracy: Is the information in the answer consistent with the reference answer?} \\
& \texttt{4. Additional information: If the answer contains more information than the reference answer, does it remain consistent and not contradict itself?} \\
& \\
& \texttt{Your response should be structured as follows:} \\
& \texttt{1. A justification for your decision, explaining your reasoning based on the evaluation criteria.} \\
& \texttt{2. A grade of either "correct" or "incorrect".} \\
& \\
& \texttt{Important note on deflections and invalid questions:} \\
& \texttt{- If the answer is a deflection or does not attempt to answer the question, grade it as "incorrect" unless the reference answer is also a deflection.} \\
& \texttt{- If both the answer and the reference answer indicate that the question is invalid or cannot be answered, grade it as "correct".} \\
& \texttt{- If the answer is a placeholder like \{\{YOUR\_ANSWER\}\}, any generic phrase that does NOT address the question, or an empty answer, then count it as "incorrect".} \\
& \\
& \texttt{Your response should be in json format as follows:} \\
& \texttt{\{"justification": "...", "grade": "correct" or "incorrect"\}} \\
\midrule
User &
\texttt{Here is the question:} \\
& \texttt{\{query\}} \\
& \\
& \texttt{Here is the answer to be judged:} \\
& \texttt{\{answer\}} \\
& \\
& \texttt{Here is the reference answer:} \\
& \texttt{\{expected\_answer\}} \\
\bottomrule
\end{tabular}
\caption{LLM-as-a-Judge prompt used for scoring free-form answers on TechQA. The judge model (DeepSeek-Distilled-Qwen-32B) receives the system prompt defining evaluation criteria and a user message containing the question, generated answer, and reference answer.}
\label{tab:llmaaj-prompt}
\end{table*}
\end{itemize}

For all datasets, we report accuracy as the percentage of correctly answered questions.

\section{Cartridge Size Statistics}
\label{sec:cartridge_sizes}

Because each cartridge is sized proportionally to its source document, the actual number of KV tokens per cartridge varies across documents within a dataset.
Table~\ref{tab:cart_sizes} reports the minimum, mean, and maximum cartridge sizes (in tokens) for each dataset at each target compression ratio used in our experiments.

All cartridge sizes are rounded up to the nearest multiple of 16 tokens.
This constraint arises from the paged KV-cache implementation in our inference server (tokasaurus\footnote{\url{https://github.com/ScalingIntelligence/tokasaurus}}), which manages memory in fixed-size pages of 16 tokens.
Aligning cartridge sizes to page boundaries avoids internal fragmentation and ensures that each cartridge occupies an integer number of pages, enabling efficient memory allocation and deallocation during both training (pool rotation) and inference (cartridge loading/unloading).
We additionally enforce a minimum cartridge size of 16 tokens (one page), ensuring that even the shortest documents receive at least one full page of KV representation.

\begin{table*}[t]
\centering
\small
\begin{tabular}{llrrrr}
\toprule
\textbf{Dataset} & \textbf{Comp.} & \textbf{Min} & \textbf{Mean} & \textbf{Max} & \textbf{Total} \\
\midrule
\multirow{6}{*}{LongHealth (20 docs)}
  & $2\times$   & 5{,}040 & 5{,}859 & 6{,}640 & 117{,}184 \\
  & $5\times$   & 2{,}016 & 2{,}349 & 2{,}656 & 46{,}976 \\
  & $10\times$  & 1{,}008 & 1{,}178 & 1{,}328 & 23{,}568 \\
  & $20\times$  & 512 & 593 & 672 & 11{,}856 \\
  & $50\times$  & 208 & 242 & 272 & 4{,}848 \\
  & $100\times$ & 112 & 126 & 144 & 2{,}512 \\
\midrule
\multirow{7}{*}{QASPER (407 docs)}
  & $2\times$   & 464 & 2{,}349 & 4{,}912 & 955{,}968 \\
  & $3\times$   & 304 & 1{,}569 & 3{,}280 & 638{,}400 \\
  & $5\times$   & 192 & 944 & 1{,}968 & 384{,}336 \\
  & $10\times$  & 96 & 476 & 992 & 193{,}904 \\
  & $20\times$  & 64 & 242 & 496 & 98{,}640 \\
  & $50\times$  & 64 & 103 & 208 & 42{,}016 \\
  & $100\times$ & 64 & 68 & 112 & 27{,}568 \\
\midrule
\multirow{6}{*}{QuALITY (115 docs)}
  & $2\times$   & 1{,}136 & 2{,}865 & 4{,}208 & 329{,}456 \\
  & $5\times$   & 464 & 1{,}151 & 1{,}696 & 132{,}336 \\
  & $10\times$  & 240 & 579 & 848 & 66{,}544 \\
  & $20\times$  & 128 & 293 & 432 & 33{,}744 \\
  & $50\times$  & 64 & 123 & 176 & 14{,}176 \\
  & $100\times$ & 64 & 74 & 96 & 8{,}464 \\
\midrule
\multirow{6}{*}{FinQA (380 docs)}
  & $2\times$   & 112 & 524 & 1{,}280 & 199{,}184 \\
  & $5\times$   & 64 & 214 & 512 & 81{,}456 \\
  & $10\times$  & 64 & 112 & 256 & 42{,}528 \\
  & $20\times$  & 64 & 69 & 128 & 26{,}096 \\
  & $50\times$  & 16 & 29 & 64 & 11{,}088 \\
  & $100\times$ & 16 & 17 & 32 & 6{,}448 \\
\midrule
\multirow{6}{*}{TechQA (471 docs)}
  & $2\times$   & 64 & 460 & 2{,}048 & 216{,}832 \\
  & $5\times$   & 64 & 193 & 1{,}008 & 90{,}784 \\
  & $10\times$  & 64 & 109 & 512 & 51{,}248 \\
  & $20\times$  & 64 & 75 & 256 & 35{,}328 \\
  & $50\times$  & 64 & 65 & 112 & 30{,}496 \\
  & $100\times$ & 16 & 19 & 64 & 8{,}752 \\
\bottomrule
\end{tabular}
\caption{Cartridge size statistics (in tokens) per dataset and target compression ratio. \textbf{Min}: smallest cartridge in the collection. \textbf{Mean}: average tokens per cartridge. \textbf{Max}: largest cartridge. \textbf{Total}: sum of all cartridge sizes. All sizes are multiples of 16 (the page size).}
\label{tab:cart_sizes}
\end{table*}

At high compression ratios ($50\times$--$100\times$), the shortest documents in FinQA and TechQA are clamped to the 16-token minimum.
For example, a 500-token FinQA document at $100\times$ compression would nominally require only 5 tokens, but is rounded up to 16.
This means the \emph{effective} compression ratio for short documents is lower than the target---a 500-token document with a 16-token cartridge achieves $31\times$ rather than $100\times$.
Conversely, the longest documents in each collection achieve compression ratios close to the target.
The wide min--max range within a dataset (e.g., 64--2{,}048 for TechQA at $2\times$) reflects the natural variation in document lengths: the proportional sizing ensures that each document receives a KV budget commensurate with its information content.

\section{RAG Baseline: Indexing and Retrieval Details}
\label{sec:rag_details}

We describe the full RAG pipeline below.

\paragraph{Document indexing.}
Each dataset's documents are indexed with chunk size $C \in \{128, 256, 512, 1024\}$ tokens.
Documents are chunked using a fixed-size strategy with a 10\% token overlap between adjacent chunks.
Chunks are embedded using Amazon Titan Embed Text v2 (\texttt{amazon.titan-embed-text-v2:0}) via Amazon Bedrock\footnote{\url{https://aws.amazon.com/bedrock}} with 1024-dimensional embeddings and retrieved via cosine similarity.

\paragraph{Retrieval.}
At evaluation time, each question is issued as a retrieval query against the Knowledge Base.
We retrieve the top-$K$ chunks for $K \in \{1, 3, 5, 10\}$.
To avoid redundant API calls, we retrieve $K = 10$ once per chunk size and slice for smaller $K$ values.
Retrieved chunks are concatenated in score order and prepended to the question as the context for the reader model.

\paragraph{Hyperparameter selection.}
We report the best result across all 16 chunk/top-$K$ combinations ($4 \times 4$) for each dataset as the ``RAG best'' entry in Table~\ref{tab:compression_ablation}.
The optimal configuration varies by dataset: datasets with short, self-contained answers (TechQA, FinQA) benefit from larger chunks and higher $K$, while datasets requiring precise span extraction (QASPER) are less sensitive to retrieval configuration.

\paragraph{Effective token counts.}
Table~\ref{tab:rag_tokens} reports the average number of context tokens consumed per query for Text RAG at chunk size 1024 across different $k$ values. Due to the 10\% overlap between adjacent chunks and shorter final chunks in some documents, the effective token count is slightly below the nominal $k \times 1024$.

\paragraph{Unique cartridges per query.}
Because multiple retrieved chunks may originate from the same source document, the number of unique documents loaded in Cartridge RAG is often lower than the number of retrieved chunks $k$.
Table~\ref{tab:unique_docs} reports the average number of unique documents covered by the top-$k$ chunks (at chunk size 1024).
Datasets with short documents (FinQA, TechQA) exhibit high chunk diversity---10 chunks span 8--9 distinct documents---while datasets with longer documents (LongHealth, QuALITY) show more within-document clustering, with 10 chunks covering only 4--5 documents on average.

\begin{table}[t]
\centering
\small
\begin{tabular}{lcccc}
\toprule
\textbf{Dataset} & $k{=}1$ & $k{=}3$ & $k{=}5$ & $k{=}10$ \\
\midrule
LongHealth & 1.0 & 1.8 & 2.5 & 4.4 \\
QuALITY    & 1.0 & 1.4 & 2.2 & 4.8 \\
FinQA      & 1.0 & 3.0 & 5.0 & 8.7 \\
TechQA     & 1.0 & 3.0 & 4.9 & 8.1 \\
\bottomrule
\end{tabular}
\caption{Average number of unique source documents covered by the top-$k$ retrieved chunks (chunk size 1024). Datasets with shorter documents (FinQA, TechQA) have higher chunk diversity, while longer-document datasets (LongHealth, QuALITY) exhibit within-document clustering.}
\label{tab:unique_docs}
\end{table}

\begin{table}[t]
\centering
\small
\begin{tabular}{lcccc}
\toprule
\textbf{Dataset} & $k{=}1$ & $k{=}3$ & $k{=}5$ & $k{=}10$ \\
\midrule
LongHealth & 986 & 2{,}960 & 4{,}944 & 9{,}860 \\
QuALITY    & 941 & 2{,}856 & 4{,}733 & 9{,}412 \\
FinQA      & ${\sim}$960 & ${\sim}$2{,}880 & ${\sim}$4{,}800 & ${\sim}$9{,}600 \\
TechQA     & ${\sim}$960 & ${\sim}$2{,}880 & ${\sim}$4{,}800 & ${\sim}$9{,}600 \\
\bottomrule
\end{tabular}
\caption{Average context tokens per query for Text RAG (chunk size 1024).}
\label{tab:rag_tokens}
\end{table}

\section{Retrieval Quality: Recall and MRR}
\label{sec:rag_recall}

To contextualize the Cartridge RAG results in \S\ref{sec:experimenta_results} \emph{Cartridge RAG}, we report document-level retrieval quality of the dense retriever used by both the Text RAG and Cartridge RAG pipelines.
For each query we retrieve $10$ chunks, deduplicate to unique source documents, and compute Recall@$K$ and MRR@$K$ over the resulting ranked document list (so $K$ counts unique documents, not chunks).
Tables~\ref{tab:rag_recall_lh}--\ref{tab:rag_recall_techqa} report these metrics for the four chunk sizes used in our sweep.

\begin{table}[t]
\centering
\small
\resizebox{\columnwidth}{!}{%
\begin{tabular}{lcccccccc}
\toprule
\textbf{Chunk} & \textbf{R@1} & \textbf{R@3} & \textbf{R@5} & \textbf{R@10} & \textbf{MRR@1} & \textbf{MRR@3} & \textbf{MRR@5} & \textbf{MRR@10} \\
\midrule
128  & 79.2 & 88.5 & 91.2 & 95.8 & 79.2 & 83.6 & 84.5 & 85.5 \\
256  & 76.5 & 87.5 & 91.2 & 95.5 & 76.5 & 81.5 & 82.5 & 83.3 \\
512  & 76.0 & 89.0 & 93.0 & 97.0 & 76.0 & 81.8 & 82.8 & 83.6 \\
1024 & 76.5 & 86.5 & 91.2 & 96.8 & 76.5 & 81.0 & 82.4 & 83.4 \\
\bottomrule
\end{tabular}%
}
\caption{LongHealth document-level retrieval ($20$ docs, $400$ queries).}
\label{tab:rag_recall_lh}
\end{table}

\begin{table}[t]
\centering
\small
\resizebox{\columnwidth}{!}{%
\begin{tabular}{lcccccccc}
\toprule
\textbf{Chunk} & \textbf{R@1} & \textbf{R@3} & \textbf{R@5} & \textbf{R@10} & \textbf{MRR@1} & \textbf{MRR@3} & \textbf{MRR@5} & \textbf{MRR@10} \\
\midrule
128  & 85.3 & 91.1 & 93.1 & 95.2 & 85.3 & 88.0 & 88.5 & 88.9 \\
256  & 88.0 & 92.6 & 94.0 & 95.4 & 88.0 & 90.1 & 90.5 & 90.8 \\
512  & 88.3 & 92.1 & 93.2 & 95.5 & 88.3 & 90.1 & 90.4 & 90.9 \\
1024 & 87.1 & 91.4 & 93.1 & 94.8 & 87.1 & 89.1 & 89.6 & 90.0 \\
\bottomrule
\end{tabular}%
}
\caption{QuALITY document-level retrieval ($115$ docs, $2{,}086$ queries).}
\label{tab:rag_recall_quality}
\end{table}

LongHealth and QuALITY have nearly saturated retrieval ($\geq 95\%$ Recall@10), so any gap between Text RAG and Oracle (Text) on these datasets is attributable to information loss in the chunked text rather than retrieval failures.

\begin{table}[t]
\centering
\small
\resizebox{\columnwidth}{!}{%
\begin{tabular}{lcccccccc}
\toprule
\textbf{Chunk} & \textbf{R@1} & \textbf{R@3} & \textbf{R@5} & \textbf{R@10} & \textbf{MRR@1} & \textbf{MRR@3} & \textbf{MRR@5} & \textbf{MRR@10} \\
\midrule
128  & 70.4 & 85.7 & 90.7 & 95.9 & 70.4 & 77.0 & 78.0 & 79.0 \\
256  & 70.5 & 87.3 & 91.8 & 96.1 & 70.5 & 78.0 & 79.0 & 79.0 \\
512  & 69.2 & 86.0 & 91.5 & 95.6 & 69.2 & 77.0 & 78.0 & 79.0 \\
1024 & 69.2 & 87.0 & 91.8 & 96.0 & 69.2 & 77.0 & 78.0 & 79.0 \\
\bottomrule
\end{tabular}%
}
\caption{FinQA document-level retrieval ($380$ docs, $1{,}147$ queries).}
\label{tab:rag_recall_finqa}
\end{table}

\begin{table}[t]
\centering
\small
\resizebox{\columnwidth}{!}{%
\begin{tabular}{lcccccccc}
\toprule
\textbf{Chunk} & \textbf{R@1} & \textbf{R@3} & \textbf{R@5} & \textbf{R@10} & \textbf{MRR@1} & \textbf{MRR@3} & \textbf{MRR@5} & \textbf{MRR@10} \\
\midrule
128  & 79.8 & 89.0 & 91.8 & 93.1 & 79.8 & 83.8 & 84.4 & 84.6 \\
256  & 80.3 & 89.7 & 92.6 & 93.9 & 80.3 & 84.5 & 85.2 & 85.4 \\
512  & 82.3 & 90.2 & 93.1 & 95.2 & 82.3 & 85.8 & 86.5 & 86.9 \\
1024 & 83.1 & 92.1 & 93.8 & 95.2 & 83.1 & 87.3 & 87.7 & 87.9 \\
\bottomrule
\end{tabular}%
}
\caption{TechQA document-level retrieval ($496$ docs, $610$ queries).}
\label{tab:rag_recall_techqa}
\end{table}

FinQA and TechQA both exhibit high retrieval quality ($\geq 95\%$ Recall@10), comparable to LongHealth and QuALITY.
FinQA documents contain distinctive company names and financial terminology that make retrieval effective despite the large corpus (380 docs).
TechQA's IBM Technotes cover distinct products and error codes, yielding strong R@1 ($80$--$83\%$) that improves with larger chunk sizes as more diagnostic context is captured per chunk.

\section{Evaluation Prompts}
\label{sec:eval_prompts}

We use task-specific system prompts during evaluation. All prompts instruct the model to wrap its final answer in \texttt{<answer>} tags. In the Cartridge setting, the document context is encoded in the trained KV cache prefix and is not included in the text prompt---the model receives only the system prompt and the user question. In the Oracle (Text) baseline, the full document text is prepended to the user message. Table~\ref{tab:eval-prompts} lists the exact prompts used.

For FinQA, we use the evaluation prompt recommended in the T$^2$-RAGBench paper~\cite{strich2026t2ragbench}.\footnote{\url{https://github.com/uhh-hcds/g4kmu-paper/tree/main/src/g4k/prompts/sys_prompts}} For QASPER\footnote{\url{https://github.com/EleutherAI/lm-evaluation-harness/tree/main/lm_eval/tasks/qasper}} and QuALITY, we adopt the prompts from previous work~\cite{eval-harness,eyuboglu2025cartridges,zweiger2026fast}. For LongHealth, we use the prompt format from the original benchmark\footnote{\url{https://github.com/HazyResearch/cartridges/blob/main/examples/benchmarks/longhealth/baseline_longhealth.py}}. For TechQA, since no standard evaluation prompt exists, we conducted a prompt search over several variants---including a minimal instruction (``Answer concisely''), a domain-specific expert framing, a chain-of-thought variant, and a structured extraction format---and selected the prompt that yielded the highest Oracle (Text) accuracy on the development set.

\begin{table*}[t]
\centering
\footnotesize
\begin{tabular}{@{}p{0.15\textwidth}p{0.82\textwidth}@{}}
\toprule
\textbf{Benchmark} & \textbf{System Prompt + User Message Format} \\
\midrule
T$^2$-RAGBench / FinQA &
\texttt{System: You are an expert in answering financial questions by constructing mathematical formulas based on a simple syntax.} \\
& \texttt{- Task: Provide a FORMULA ANSWER to the question based on the given context.} \\
& \texttt{Guidelines:} \\
& \texttt{1. Answer Type: A formula is either a number or one of: add(f1, f2), subtract(f1, f2), multiply(f1, f2), divide(f1, f2), exp(f1, f2), greater(f1, f2)} \\
& \texttt{2. Reasoning: Carefully analyze the context. Pay special attention to the table.} \\
& \texttt{3. Final Answer: "<answer> FORMULA </answer>"} \\
& \\
& \texttt{User: \{context\}} \\
& \texttt{Question: \{question\}} \\
\midrule
LongHealth &
\texttt{System: Please reference the patient medical records to answer the user's questions. Choose the single best option and provide your answer exactly as it appears in the options.} \\
& \texttt{Wrap your answer in: <answer> The correct option text here </answer>} \\
& \\
& \texttt{User: \{question\}} \\
& \texttt{A. \{option\_a\}~~B. \{option\_b\}~~C. \{option\_c\}~~D. \{option\_d\}~~E. \{option\_e\}} \\
\midrule
QASPER &
\texttt{System: You are a research assistant answering questions about a scientific paper. Answer as briefly as possible. Give only the answer, no explanation. If the question cannot be answered from the paper, say "Unanswerable". For yes/no questions, answer "Yes" or "No". Wrap your answer in: <answer> ... </answer>} \\
& \\
& \texttt{User: \{question\}} \\
\midrule
QuALITY &
\texttt{System: You are a careful reader answering multiple-choice questions about a long article. Read the article and choose the single best answer option. Provide your answer as the letter (A, B, C, or D) wrapped in answer tags. Example: <answer> B </answer>} \\
& \\
& \texttt{User: \{question\}} \\
& \texttt{A. \{option\_a\}~~B. \{option\_b\}~~C. \{option\_c\}~~D. \{option\_d\}} \\
& \texttt{Answer with the letter of the correct option (A, B, C, or D).} \\
\midrule
TechQA &
\texttt{System: You are an expert technical support assistant specializing in IT infrastructure, software products, and enterprise systems. Answer the technical question to the best of your ability. Be concise and factual. Wrap your answer in: <answer> YOUR\_ANSWER </answer>} \\
& \\
& \texttt{User: \{question\}} \\
\bottomrule
\end{tabular}
\caption{Evaluation prompts used for each benchmark. Variables in braces are replaced with benchmark data at evaluation time.}
\label{tab:eval-prompts}
\end{table*}

\section{Self-Study Data Synthesis Details}
\label{sec:synthesis_details}

\paragraph{Question generation model.}
We use GPT-OSS 120B~\cite{agarwal2025gpt} as the question generator ($M_Q$).
This model receives the full document context in its system prompt and generates batches of 20 diverse questions per call.
The answer generator ($M_A$) is the target model itself (Qwen3-8B), ensuring that the distillation signal reflects the student's own output distribution.

\paragraph{Seed prompt types.}
Each synthesis call randomly selects a seed prompt type from five categories: \emph{structuring} (requests to organize information into JSON, YAML, or other formats), \emph{summarization} (requests to summarize specific sections), \emph{question} (factual recall and reasoning questions), \emph{use\_case} (practical downstream application tasks), and \emph{creative} (open-ended discussion prompts).
This diversity ensures the cartridge is trained on varied interaction patterns rather than only factoid QA.

\paragraph{Multi-question prompt format.}
In batched mode, the question generator receives the following instruction (shown for the \emph{question} type):

\begin{quote}
\small
\texttt{Generate \{n\} diverse questions that test knowledge of the information in the corpus above. Each question should cover a different fact, detail, or aspect of the corpus. Vary the style: mix factual recall, comparison, reasoning, and detail-oriented questions. Include specific details (ids, names, titles, dates, numerical values, etc.) in each question so it is clear what you are asking about.}

\texttt{Output ONLY a JSON array of strings, e.g.\ ["question 1", "question 2", ...]. No other text, no markdown fences, no explanation.}
\end{quote}

The structured JSON output format enables reliable parsing; responses that fail to parse are discarded (typically ${<}5\%$ of calls).

\paragraph{Sampling rounds and temperatures.}
For each dataset, we run 4 independent sampling rounds, one per seed prompt type (question, structuring, summarization, use\_case), each generating 10{,}000 samples for a total of 40{,}000 training examples per dataset.\footnote{We exclude the \emph{creative} seed type from training data as it produces open-ended prompts less suited for distillation.}
$M_Q$ (GPT-OSS 120B, the question generator) uses temperature $0.6$ with top-$p{=}0.95$, top-$k{=}20$, and a maximum of 4{,}096 completion tokens per call.
$M_A$ (Qwen3-8B, the teacher/answer generator) uses temperature $0.0$ (greedy decoding) with a maximum of 2{,}048 completion tokens, ensuring deterministic distillation targets.
Both models operate with thinking mode enabled.

\paragraph{Sampling configuration.}
Within each round, we generate 10{,}000 synthesis samples with a batch size of 4 contexts and up to 128 parallel API calls.
Documents are sampled using the proportional-to-length strategy described in \S\ref{sec:data_synthesis}, with a fixed random seed for reproducibility.
For multi-note documents (e.g., LongHealth patient records), we sample one note per prompt to ensure fine-grained coverage of individual clinical notes.

\section{Self-Study Data Quality Analysis}
\label{sec:synthesis_analysis}

To understand the failure modes of the original Self-Study procedure, we conduct a detailed analysis on 9 FinQA cartridges (512 tokens each, trained for 80 epochs).
We measure \emph{value coverage}: for each unique number in the original document, we count the fraction of training samples that reproduce it.

\paragraph{Synthesizer bias toward narrative over tables.}
The synthesizer preferentially generates questions about prose commentary rather than specific table cell values.
Table~\ref{tab:synth_grounding} shows that across all 9 cartridges, only 41\% of unique numbers in the training data are grounded in the original document.
The remaining 59\% are derived calculations (62\%), rounded/reformatted values (22\%), or pure fabrications (15\%).

\begin{table}[h]
\centering
\small
\begin{tabular}{lrr}
\toprule
\textbf{Category} & \textbf{Count} & \textbf{\%} \\
\midrule
Grounded (in document) & 281 & 41\% \\
Derived calculations & 255 & 37\% \\
Rounded/reformatted & 90 & 13\% \\
Pure fabrications & 62 & 9\% \\
\midrule
Total not in document & 407 & 59\% \\
\bottomrule
\end{tabular}
\caption{Classification of unique numbers appearing in Self-Study training data across 9 FinQA cartridges (688 unique numbers total, excluding years).}
\label{tab:synth_grounding}
\end{table}

For example, in a Lockheed Martin document, the text-derived delta ``operating profit decreased \$264 million'' achieves 87\% coverage (80/92 samples), while the underlying table values it was computed from---\$6{,}608M, \$6{,}770M, \$7{,}092M (net sales 2014--2016)---appear at only 3--5\% coverage.
All 4 test questions requiring these table values score 0\%.

\paragraph{Does coverage predicts correctness?}
Table~\ref{tab:coverage_threshold} shows the relationship between minimum value coverage and answer correctness.
A coverage threshold of ${\sim}25\%$ is necessary (87\% of correct answers meet it) but not sufficient (46\% of incorrect answers also meet it), indicating that additional failure modes---such as entity confusion---contribute to errors even when values are present.

\begin{table}[h]
\centering
\small
\begin{tabular}{lcc}
\toprule
\textbf{Min coverage} & \textbf{Correct ($n{=}15$)} & \textbf{Incorrect ($n{=}13$)} \\
\midrule
$\geq$ 0\% & 100\% & 100\% \\
$\geq$ 10\% & 100\% & 69\% \\
$\geq$ 25\% & 87\% & 46\% \\
$\geq$ 30\% & 67\% & 23\% \\
$\geq$ 50\% & 27\% & 23\% \\
\bottomrule
\end{tabular}
\caption{Fraction of questions meeting a minimum coverage threshold, split by correctness. Coverage $\geq 25\%$ is necessary but not sufficient.}
\label{tab:coverage_threshold}
\end{table}

\paragraph{Entity confusion.}
The cartridge stores numeric values but loses entity-value associations.
In a UPS shareowner return document, the model correctly retrieves UPS's return (\$121.46, 100\% coverage) but substitutes the S\&P~500 value (\$108.59, 69\% coverage) for the Dow Jones Transportation value (\$127.07, 69\% coverage) when computing their difference---both distractor values have identical coverage, and the model picks the wrong entity.

\paragraph{Hallucination despite self-verification.}
The model fabricates plausible values, attempts chain-of-thought verification, detects inconsistencies, and proceeds anyway.
In a Hologic acquisition document, the model needs the goodwill value (\$6{,}900, 28\% coverage) but produces three different hallucinated values across 3 runs (4{,}200, 6{,}000, 3{,}000).
In each run, it sums the allocation components, notices the total does not match \$31{,}300, yet outputs the wrong formula regardless.

\section{Artifact Licenses and Terms of Use}
\label{appx:licenses}

\paragraph{Licenses and terms for use/distribution of artifacts.}

The five benchmarks used in this work are publicly available under the following licenses:

\begin{itemize}[nosep]
    \item \textbf{LongHealth}~\cite{adams2024longhealth}: Released under the Apache-2.0 License.\footnote{\url{https://github.com/kbressem/LongHealth/blob/main/LICENSE}} The dataset consists of entirely fictional patient records created by the authors; no real patient data is included.
    \item \textbf{QASPER}~\cite{dasigi2021qasper}: Released under the CC BY 4.0 License.\footnote{\url{https://huggingface.co/datasets/allenai/qasper}} The dataset consists of questions and answers over NLP research papers; the dataset itself is distributed under CC BY 4.0.
    \item \textbf{QuALITY}~\cite{pang2022quality}: Released under the CC BY 4.0 License.\footnote{\url{https://nyu-mll.github.io/quality/}} Source texts are drawn from Project Gutenberg (public domain) and other permissively licensed collections, including nonfiction and fiction sources such as Slate articles from the Open American National Corpus, The Long+Short, Freesouls, and Open Access books. These texts are published works and do not contain personal data, though some may include mature themes typical of literary and journalistic content.
    \item \textbf{FinQA}~\cite{chen2021finqa}: Released under the MIT License.\footnote{\url{https://github.com/czyssrs/FinQA/blob/main/LICENSE}} The underlying financial reports are sourced from corporate filings made available via the SEC EDGAR system, including earnings reports and annual/quarterly reports (e.g., 10-K and 10-Q documents). These are publicly accessible financial disclosures prepared by reporting companies. T$^2$-RAGBench~\cite{strich2026t2ragbench} is also released under the MIT License.
    \item \textbf{TechQA}~\cite{castelli2020techqa}: The NVIDIA TechQA-RAG-Eval variant used in this work is released under the Apache-2.0 License and is explicitly cleared for commercial and non-commercial use.\footnote{\url{https://huggingface.co/datasets/nvidia/TechQA-RAG-Eval}} It is derived from the original IBM TechQA dataset, whose code repository is also Apache-2.0.\footnote{\url{https://github.com/IBM/techqa/blob/master/LICENSE.md}}
\end{itemize}

The base model used in our experiments, \textbf{Qwen3-8B}~\cite{yang2025qwen3}, is released under the Apache 2.0 License, which permits research and commercial use.

We do not release new datasets in this work. The trained cartridge artifacts (KV cache parameters) are derivatives of the above datasets and the Qwen3-8B model weights; any release of such artifacts would be subject to the intersection of the applicable licenses above.

\paragraph{ Offensive content and personal data.}

\begin{itemize}[nosep]
    \item \textbf{LongHealth}: The dataset consists of entirely fictional patient records with no real individuals. No anonymization was required. We verified that no real names, addresses, or identifying information appear in the data.
    \item \textbf{QASPER}: Source texts are NLP research papers. No personal data or offensive content is expected; the domain is scientific writing.
    \item \textbf{QuALITY}: Source texts are fiction and non-fiction narratives, these are published literary works; no personal data is present.
    \item \textbf{FinQA / T$^2$-RAGBench}: Source texts are corporate earnings reports from SEC EDGAR. These are formal financial documents; no personal data or offensive content is present.
    \item \textbf{TechQA}: Source texts are IBM Technote IT support documents. These are technical documentation; no personal data or offensive content is expected.
\end{itemize}

No additional anonymization steps were taken beyond those applied by the original dataset creators, as none of the datasets contain personal data about private individuals.

\section{AI use disclosure} We used AI to assist with code writing and manuscript typesetting. 

\end{document}